\documentclass{article}

\usepackage[nonatbib,preprint]{neurips_2019}

\usepackage[utf8]{inputenc} 
\usepackage[T1]{fontenc}    
\usepackage{hyperref}       
\usepackage{url}            
\usepackage{booktabs}       
\usepackage{amsfonts}       
\usepackage{nicefrac}       
\usepackage{microtype}      
\usepackage{amsmath}        
\usepackage{math_commands}  
\usepackage{textgreek}      
\usepackage{algorithm}      
\usepackage{algpseudocode} 
\usepackage{multirow}
\usepackage{graphicx}
\usepackage[colorinlistoftodos]{todonotes} 

\usepackage{parskip}

\title{Is Q-Learning Provably Efficient? \\ An Extended Analysis}

\author{%
    Kushagra Rastogi \thanks{Indicates equal contribution.} \\
    University of California, Los Angeles \\
    Los Angeles, CA 90095 \\
    \texttt{krastogi@g.ucla.edu}
    \And
    Jonathan Lee \footnotemark[1] \\
    University of California, Los Angeles \\
    Los Angeles, CA 90095 \\
    \texttt{jlee916@g.ucla.edu} 
    \And
    Fabrice Harel-Canada \footnotemark[1] \\
    University of California, Los Angeles \\
    Los Angeles, CA 90095 \\
    \texttt{fabricehc@cs.ucla.edu}
    \And   
    Aditya Joglekar \\
    University of California, Los Angeles \\
    Los Angeles, CA 90095 \\
    \texttt{adivj123@gmail.com}
}

\begin{document}

\maketitle

\begin{abstract}
This work extends the analysis of the theoretical results presented within the paper \emph{Is Q-Learning Provably Efficient?} by Jin \emph{et al.} \cite{Jin2018IsQP}. We include a survey of related research to contextualize the need for strengthening the theoretical guarantees related to perhaps the most important threads of model-free reinforcement learning. We also expound upon the reasoning used in the proofs to highlight the critical steps leading to the main result showing that Q-learning with UCB exploration achieves a sample efficiency that matches the optimal regret that can be achieved by any model-based approach.  
\end{abstract}

\section*{Introduction}

State-of-the-art reinforcement learning (RL) has been dominated by model-free algorithms (like Q-learning) because they are online, more expressive and need less space. However, empirical work has shown that model-free algorithms have a higher sample complexity \cite{PILCO, schulman2015trust}, meaning that they require many more samples in order to perform well on a given task. Can we make model-free algorithms sample-efficient? This is one of the most fundamental questions in the reinforcement learning community that has yet to be answered definitely. As seen in the setting of multi-armed bandits, good sample efficiency is the result of aptly managing the exploration-exploitation trade-off. In our project, we aim to elaborate on the proofs establishing that Q-learning with Upper Confidence Bound (UCB) exploration, in an episodic MDP setting and without access to a ``simulator'', matches the information-theoretic regret optimum, up to a single $\sqrt{H}$ where \textit{H} is the number of steps per episode. To do this, we will leverage our current understanding of Q-learning and survey existing literature related to sample efficiency and complexity of both model-free and model-based RL methods. 
\section*{Related Work}

This section reviews related work that compares Model-free (MF) and Model-based (MB) reinforcement learning (RL) in general before focusing on theoretical research into their respective sample efficiencies and complexities. 

\subsection*{Model-free vs. Model-based RL}

The study of reinforcement learning has given rise to two primary approaches for maximizing cumulative rewards while interacting with an unknown environment through time: model-based and model-free algorithms. MB algorithms are the ``planners'' that either learn or use a model of environmental dynamics to form a form a suitable control policy. On the other hand, MF algorithms make no attempt to model state transitions explicitly, instead updating their state and action value functions directly. Both fundamentally and in practice, the two approaches overlap substantially; indeed MF methods act as important building blocks for MB methods \cite{Sutton1998}. 

Despite the similarities, MF methods like classical Q-learning \cite{Watkins:1989-QLearning}, DQNs \cite{DQN} and their variants \cite{DoubleQLearning, DQN_prioritized_replay}, most policy gradient approaches \cite{PolicyGradient, A3C, schulman2015trust}, and many others dominate most of the progress in modern RL \cite{Jin2018IsQP}. Table \ref{tab:MF_MB_pros_and_cons} highlights some of the pros and cons of both approaches and highlights why MF methods enjoy wide attention in the field. Of the cons, the most problematic is the tendency for MF approaches to be sample inefficient as they require many ``experiences'' to train. The current work we analyze by Jin \emph{et al.} \cite{Jin2018IsQP} establishes that this con does not apply to the entire class of MF algorithms by showing that not only is it possible to design MF algorithms that are sample efficient, but also that Q-learning with an upper confidence bound (UCB) exploration policy \emph{is provably efficient}. However, before expanding on the illustrative process and proofs, we review other work related to sample efficiency and complexity in the next subsection. 

\begin{table}[!htbp]
    \centering
    \resizebox{\textwidth}{!}{%
    \begin{tabular}{c c c}
        \toprule
        \hfill & \textbf{Model-free (MF)}  & \textbf{Model-based (MB)}  \\
        \midrule
        \textbf{Pros} & 
        \begin{tabular}{p{0.45\textwidth}} 
            $\circ$ Computationally less complex than MB methods, requiring no model of the environment to be effective (which can be a bottleneck for MB methods) \cite{Sutton1998} \\
            $\circ$ Capable of functioning online (as opposed to working with batches) \cite{Jin2018IsQP}\\
            $\circ$ Require less space (memory) \cite{Jin2018IsQP} \\
            $\circ$ More expressive since specifying value functions / policies are more flexible than specifying a model for the environment \cite{Jin2018IsQP} \\
        \end{tabular} 
        &     
        \begin{tabular}{p{0.45\textwidth}} 
            $\circ$ Tend to be more sample efficient \cite{PILCO, Atkeson97-MB-MF-comparison} \\
            $\circ$ More efficient handling of changing goals because it does not need ``personal experience'' with every state-action pair \cite{Atkeson97-MB-MF-comparison, Sutton1998} \\ 
        \end{tabular} \\
        \hline
        \textbf{Cons} & 
        \begin{tabular}{p{0.45\textwidth}} 
            $\circ$ Requires (repeated) ``personal experience'' with many state-action pairs in order to train, makes exploration more costly \cite{Sutton1998} \\
            $\circ$ Tend to be less sample efficient \cite{Sutton1998, PILCO, Atkeson97-MB-MF-comparison} \\
        \end{tabular} 
        &         
        \begin{tabular}{p{0.45\textwidth}} 
            $\circ$ Suffer from model bias, i.e., they inherently assume that the learned dynamics model sufficiently accurately resembles the real environment \cite{PILCO, Schneider-MB-bias, Schaal-MB-bias, Atkeson97-MB-MF-comparison} \\
            $\circ$ Computationally more complex than MF methods - can be difficult to learn a good model of state transitions / rewards \cite{Sutton1998} \\
        \end{tabular} \\
        \bottomrule
    \end{tabular}%
    }
    \caption{Pros \& Cons of MF vs. MB RL Approaches}
    \label{tab:MF_MB_pros_and_cons}
    \vspace{-6mm}
\end{table}

\subsection*{Sample Efficiency \& Complexity}

Within RL, \emph{sample efficiency} $e(\cdot)$ measures the number of inputs an agent requires in order to achieve a given level of performance \cite{everitt2002cambridge} on a particular task. For example, for any two agents $A_1$ and $A_2$, $e(A_1) > e(A_2)$ if $A_1$ requires fewer inputs to achieve the \emph{same} performance as $A_2$ on a given task. The related idea of \emph{sample complexity} measures the minimum number of inputs required to guarantee a probably approximately correct (PAC) estimator \cite{PAC}. Generally, the lower the sample complexity, the more efficient the class of estimators / agents. 

In the MF setting, several recent works provide empirical evidence that MF algorithms generally require higher sample complexity \cite{PILCO, schulman2015trust}. In these cases, the authors elected to measure the duration of interactions between the agent and the environment rather than the more literal count of inputs since there is a one-to-one correspondence between the two units of measure. As an illustrative example, the authors of PILCO \cite{PILCO} measure their MB approach against six MF approaches \cite{Kimura, doya, Coulom-2002a, Wawrzynski, Riedmiller, Raiko, VanHasselt:2011} and achieved up to 5x orders of magnitude reduction in time required to succeed at the classic \emph{cart-pole} task. 

In the MB setting, several publications \cite{Agrawal2017OptimisticPS, Azar2017MinimaxRB, UCRL2, vUCQ, Osband2016GeneralizationAE} have been able to demonstrate asymptotically optimal sample efficiency by importing ideas from the bandit literature, such as the UCB variations that our selected paper also pairs with Q-learning to prove its efficiency. If the existence of a simulator is assumed, MF methods like Speedy Q-Learning \cite{SpeedyQ} can be \emph{almost} as efficient as the best MB algorithms \cite{Gheshlaghi}. Unfortunately, the value of this work is undercut by the observation that simulators generally do not do a good job of representing real-world environments where exploration is significantly harder --- i.e. using a uniformly random exploration policy is optimal for the simulator in question \cite{Gheshlaghi}. The only theoretical result for MF without using a simulator is that of ``delayed Q-learning''  by Strehl \emph{et al.} \cite{PAC-RL}, which achieves a total regret of $\cO(T^{4/5})$ --- ignoring factors in $S$, $A$, and $H$ --- compared to $\cO(\sqrt{T})$ achieved by MB methods. 

This general issue with MF methods suggests that it may be fruitful to combine key elements of MF and MB approaches to increase sample efficiency. While there is presently no theoretical basis for the benefits of this line of inquiry, several researchers have \cite{Nagabandi-NND, Pong2018TemporalDM} have demonstrated that there is at least some empirical evidence supporting the utility of blending both approaches. Nagabandi \emph{et al.} \cite{Nagabandi-NND} combine the expressiveness of deep neural networks with a model-based controller (MBC) to achieve $3-5 \times$ efficiency improvement over MF baselines on the MuJoCo \cite{MuJoCo} locomotion benchmark. Similarly, Pong \emph{et al.} \cite{Pong2018TemporalDM} proposed the ideas of temporal difference models (TDMs), which are a family of goal-conditioned value functions trained with MF learning, but used for MB control. Their experimental results show substantial improvements in efficiency relative to \emph{both} high performing MF methods like DDPG \cite{Lillicrap2015-DDPG} and HER \cite{HER} as well as MB methods on a range of RL tasks. 

Table \ref{tab:other_bounds} summarizes the regret of various algorithms discussed above and illustrates the comparative sample efficiency of the work done in our selected paper by Jin \emph{et al.} \cite{Jin2018IsQP}.

\begin{table}[!htbp]
    \centering
    \begin{tabular}{|c|c|c|c|c|}
    \hline
    \hfill & \textbf{Algorithm} & \textbf{Regret} & \textbf{Time} & \textbf{Space} \\
    \hline
    \multicolumn{1}{|c|}{\multirow{4}{*}{\textbf{MB}}} & UCRL2 \cite{UCRL2} & $\geq \cO(\sqrt{H^4S^2AT})$ & \multirow{2}{*}{$\Omega(TS^2A)$} & \multirow{4}{*}{$\cO(S^2AH)$} \\ \cline{2-3}
    \multicolumn{1}{|c|}{} & Agrawal \& Jia \cite{Agrawal2017OptimisticPS} & $\geq \cO(\sqrt{H^3S^2AT})$ &  &  \\ \cline{2-4}
    \multicolumn{1}{|c|}{} & UCBVI \cite{Azar2017MinimaxRB} & $\cO(\sqrt{H^2SAT})$ & \multirow{2}{*}{$\cO(TS^2A)$} &  \\ \cline{2-3}
    \multicolumn{1}{|c|}{} & vUCQ \cite{vUCQ} & $\cO(\sqrt{H^2SAT})$ &  &  \\ 
    \hline
    \multirow{3}{*}{\textbf{MF}}  & Delayed Q-learning \cite{PAC-RL} & $\cO_{S,A,H}(T^{4/5})$ & \multirow{3}{*}{$\cO(T)$} & \multirow{3}{*}{$\cO(SAH)$} \\ \cline{2-3}
     & Q-learning (UCB-H) \cite{Jin2018IsQP} & $\cO(\sqrt{H^4SAT})$ &  &  \\ \cline{2-3}
     & Q-learning (UCB-B) \cite{Jin2018IsQP} & $\cO(\sqrt{H^3SAT})$ &  &  \\ \hline
     & information theoretic lower bound \cite{Jin2018IsQP} & $\Omega(\sqrt{H^2SAT})$ & --- & --- \\ 
     \hline
    \end{tabular}
    \caption{Regret comparisons for RL methods on Episodic MDP where $T = KH$ is the total number of steps, $H$ is the steps per episode, $S$ is the number of states, and $A$ is the number of actions. NOTE: this table is presented for $T \geq \text{poly}(S,A,H)$, and thus omits the lower order terms.}
    \label{tab:other_bounds}
    \vspace{-6mm}
\end{table}
\section*{Preliminary}

The notation used in this paper is mostly adapted from \cite{Jin2018IsQP}.We consider an episodic Markov Decision Process (MDP) $\cM = (\cS, \cA, H, \PP, r)$, where $\cS$ is a finite set of states with $|\cS| = S$, $\cA$ is a finite set of actions with $|\cA| = A$, $H$ is the number of steps in each episode, $\PP$ is the transition matrix where $\PP_h(\cdot| x,a)$ is the distribution of states when action $a$ is taken at state $x$ at step $h\in[H]$ and $r_h: \cS\times\cA \rightarrow [0,1]$ is a deterministic reward function at step $h$.

Each episode of the MDP begins with the agent at state $x_1$. For each step $h\in[H]$, the agent observes state $x_h \in \cS$, takes action $a_h \in \cA$, receives reward $r_h(x_h,a_h)$ and subsequently transitions to the next state $x_{h+1}$ that is drawn from $\PP_h(\cdot|x_h,a_h)$. The episode ends when the agent reaches the terminal state $x_{H+1}$.

We define $V^\pi_h:\cS\rightarrow\RR$ as the agent's state-value function at step $h$ under policy $\pi$. We define $Q^\pi_h:\cS\times\cA\rightarrow\RR$ as the agent's Q-value function at step $h$ under policy $\pi$.
\begin{equation*}
    V_h^\pi(x)=\EE\bigg[\sum_{h'=h}^Hr_{h'}(x_{h'},\pi_{h'}(x_{h'})) | x_h=x\bigg]
\end{equation*}
\begin{equation*}
    Q_h^\pi(x,a)=r_h(x,a)+\EE\bigg[\sum_{h'=h+1}^Hr_{h'}(x_{h'},\pi_{h'}(x_{h'})) | x_h=x,a_h=a\bigg]
\end{equation*}

For finite state and action spaces, we define the optimal state-value function as $V^*_h(x)=\max_{\pi}V^\pi_h$ $\forall x\in\cS$ and $h\in[H]$ with optimal policy $\pi^*$. Let the total number of episodes be $K$, initial state be $x^k_1$ for episode $k$ and policy be $\pi_k$ for the $k$th episode. Then, the total expected regret is $Regret(K) = \sum^{K}_{k=1}[V^*_1(x^k_1)-V^{\pi_k}_1(x^k_1)]$.
\section*{Main Results}


We combine Q-learning with a UCB exploration strategy which has the following Q-value update: $Q_h(x,a) \gets (1-\alpha_t)Q_h(x,a)+\alpha_t[r_h(x,a)+V_{h+1}(x')+b_t]$ where $t$ counts the number of times the algorithm has visited state-action pair $(x,a)$ at step $h$, $x'$ is the next state, $b_t$ is the confidence bonus and $\alpha_t = \frac{H+1}{H+t}$ is the step-size (learning rate). This choice of $\alpha_t$ scales as $\cO(H/t)$ which allows the regret to be sub-exponential in $H$, thus making Q-learning efficient.

\textbf{Q-learning with Hoeffding bonus.} Since $r_h\in[0,1]$ and there are $H$ steps in each episode, the Q-values are upper-bounded by $H$. By the Azuma-Hoeffding inequality, the Q-values confidence bound scales as $\cO(1/\sqrt{t})$ if the state-action pair $(x,a)$ is visited $t$ times. Thus, a simple bonus would be $b_t = \cO\bigg(\sqrt{\frac{H^3\iota}{t}}\bigg)$ where $\iota = \log(SAT/p)$. We present Q-learning algorithm with UCB-Hoeffding bonus.

\begin{algorithm}
\caption{Q-learning with UCB-Hoeffding}
\begin{algorithmic}[1]
\State Initialize $Q_h(x,a) \gets H$, $N_h(x,a) \gets 0$  $\forall (x,a,h) \in \cS\times\cA\times[H]$
\For{episode $k = 1$ to $K$}
    \State get $x_1$
    \For{step $h = 1$ to $H$}
        \State $a_h \gets argmax_{a'} Q_h(x_h,a')$
        \State $t = N_h(x,a) \gets N_h(x,a) + 1$
        \State $b_t \gets c\sqrt{H^3\iota/t}$ where $c>0$ is a constant and $\iota = \log(SAT/p)$
        \State $Q_h(x_h,a_h) \gets (1-\alpha_t)Q_h(x_h,a_h)+\alpha_t[r_h(x_h,a_h)+V_{h+1}(x_{h+1})+b_t]$
        \State $V_h(x_h)\gets \min\bigg(H, \max_{a'\in\cA}Q_h(x_h,a')\bigg)$
    \EndFor
\EndFor
\end{algorithmic}
\end{algorithm}

\textbf{Theorem 1} (Hoeffding). \textit{If $b_t=c\sqrt{H^3\iota/t}$, then with probability $1-p$ $\forall p \in (0,1)$, the total regret of Algorithm 1 is at most $\cO(\sqrt{H^4SAT\iota})$ where $c>0$ is a constant and $\iota = \log(SAT/p)$.}

Algorithm 1 has a $\sqrt{T}$ regret without having access to a simulator which makes it very efficient and comparable to model-based algorithms. As an online learning algorithm, Algorithm 1 only stores the Q-value table and has superior time and space complexities when $|\cS|$ is large.

\textbf{Theorem 2} (Bernstein). \textit{For a specified $b_t$, with probability $1-p$ $\forall p \in (0,1)$, the total regret of Q-learning with UCB-Bernstein exploration is at most $\cO(\sqrt{H^3SAT\iota}+\sqrt{H^9S^3A^3\iota^4})$.}

Q-learning with UCB-Bernstein exploration improves the total regret by a factor of $\sqrt{H}$ over Q-learning with UCB-Hoeffding exploration. Thus, the asymptotic regret of UCB-Bernstein is only a $\sqrt{H}$ factor away from the optimal regret achieved by model-based algorithms. However, when $T$ is small, total regret of UCB-Bernstein exploration is dominated by $\cO(\sqrt{H^9S^3A^3\iota^4})$.

\textbf{Theorem 3} (Information-theoretic lower bound). \textit{The total regret for any algorithm in an episodic MDP setting must be at least $\Omega(\sqrt{H^2SAT})$.} 

Note that the upper bounds mentioned in Theorem 1 and 2 differ from the optimal regret by a factor of $H$ and $\sqrt{H}$ respectively.
\section*{Proofs for Algorithm 1}

\textbf{Notation.} We have $(x^k_h,a^k_h)=$ the state-action pair observed and chosen at step $h$ of episode $k$. $\II[A]$ is the indicator function for event $A$. We use $Q_h^k, V_h^k, N_h^k$ to represent the $Q_h, V_h, N_h$ functions at the beginning of episode $k$. We get the following update rules for Algorithm 1:
\begin{equation*}
    V_h^k(x) \gets \min\bigg(H,\max_{a'\in\cA} Q_h^k(x,a')\bigg), \forall x\in\cS
\end{equation*}
\begin{equation}\label{eq:1}
    Q_h^{k+1}(x,a)=
    \begin{cases}
    (1-\alpha_t)Q_h^k(x,a)+\alpha_t[r_h(x,a)+V^k_{h+1}(x^k_{h+1})+b_t], & \text{if$(x,a)=(x_h^k,a_h^k)$}   \\
    Q^k_h(x,a), & \text{otherwise}
    \end{cases}
\end{equation}
We have $[\PP_hV_{h+1}](x,a)=\EE_{x'\sim\PP_h(\cdot|x,a)}V_{h+1}(x')$ and its empirical counterpart of episode $k$ is $[\hat{\PP}_h^kV_{h+1}](x,a)=V_{h+1}(x^k_{h+1})$ which is only defined for $(x,a)=(x_h^k,a_h^k)$.

The learning rate is $\alpha_t=\frac{H+1}{H+t}$. Also, we present $\alpha_t^0=\prod_{j=1}^{t}1-\alpha_j$ and $\alpha_t^i=\prod_{j=i+1}^{t}1-\alpha_j$. Since empty products are equal to 1 and empty summations equal to 0, we get $\alpha^0_t=1$ and $\sum_{i=1}^{t}\alpha_t^i=0$ for $t=0$. For $t\ge1$, we get $\alpha^0_t=\prod_{j=1}^t\frac{j-1}{H+j}=0$ and $\sum_{i=1}^{t}\alpha_t^i=1$. Combining these equations with \eqref{eq:1}, we get:
\begin{equation}\label{eq:2}
    Q_h^k(x,a)=\alpha_t^0H+\sum_{i=1}^{t}\alpha_t^i\bigg[r_h(x,a)+V^{k_i}_{h+1}(x^{k_i}_{h+1})+b_i\bigg]
\end{equation}

\textbf{Lemma 1.1.} \textit{Properties of $\alpha^i_t$:}
\begin{itemize}
    \item[\textbf{(a)}] \textit{For every $t\ge1$, $\frac{1}{\sqrt{t}}\le\sum_{i=1}^{t}\frac{\alpha_t^i}{\sqrt{i}}\le\frac{2}{\sqrt{t}}$.}
    \item[\textbf{(b)}] \textit{For every $t\ge1$, $\max_{i\in[t]}\alpha_t^i\le\frac{2H}{t}$ and $\sum_{i=1}^t(\alpha_t^i)^2\le\frac{2H}{t}$.}
    \item[\textbf{(c)}] \textit{For every $i\ge1$, $\sum_{t=i}^\infty\alpha_t^i=1+\frac{1}{H}$.}
\end{itemize}

\textit{Proof of Lemma 1.1.} Our choice of the learning rate is crucial for Q-learning to be efficient. Property (c) is particularly important to bound the regret by a constant factor of $(1+\frac{1}{H})^H$ for each step in each episode. We provide proofs for the properties.

\textbf{(a)} We use induction on $t$. For the base case $t=1$, we get $\alpha_1^1=1$. Note that $\alpha_t^i=(1-\alpha_t)\alpha_{t-1}^i$ for $i=1,...,t-1$ and $t\ge2$. This means $\sum_{i=1}^t\frac{\alpha_t^i}{\sqrt{i}}=\frac{\alpha_t}{\sqrt{t}}+(1-\alpha_t)\sum_{i=1}^{t-1}\frac{\alpha_{t-1}^i}{\sqrt{i}}$. Recall that $H\ge1$ for Q-learning to be meaningful. Using induction on both sides, we can show that $\frac{\alpha_t}{\sqrt{t}}+(1-\alpha_t)\sum_{i=1}^{t-1}\frac{\alpha_{t-1}^i}{\sqrt{i}}\ge\frac{1}{\sqrt{t}}$ and $\frac{\alpha_t}{\sqrt{t}}+(1-\alpha_t)\sum_{i=1}^{t-1}\frac{\alpha_{t-1}^i}{\sqrt{i}}\le\frac{2}{\sqrt{t}}$.

\textbf{(b)} We have $\alpha_t^i=\frac{H+1}{H+i}\bigg(\frac{i}{H+i+1}\frac{i+1}{H+i+2}...\frac{t-1}{H+t}\bigg)$. By rearranging, we get $\alpha_t^i=\frac{H+1}{H+t}\prod_{i=1}^t\frac{i}{H+i}=\max_{i\in[t]}\alpha_t^i$. Each term in the product resembles $\frac{x}{x+y}$ with $y\ge1$. Thus, $\frac{x}{x+y}\le1$ and hence $\alpha_t^i\le\frac{H+1}{H+t}$. Since, $\frac{H+1}{H+t}\le\frac{H+H}{H+t}\le\frac{H+H}{t}$, then  $\alpha_t^i\le\frac{2H}{t}$. Thus, we have shown that $\max_{i\in[t]}\alpha_t^i\le\frac{2H}{t}$. But $\sum_{i=1}^t(\alpha_t^i)(\alpha_t^i)\le\sum_{i=1}^t\alpha_t^i(\max_{i\in[t]}\alpha_t^i)$ which implies $\sum_{i=1}^t(\alpha_t^i)^2\le\frac{2H}{t}$.

\textbf{(c)} We have 
\begin{equation*}
    \begin{split}
        \sum_{t=1}^\infty\alpha_t^i=\sum_{t=1}^\infty\alpha_i\prod_{j=i+1}^t(1-\alpha_j)&=\alpha_i\sum_{t=1}^\infty\prod_{j=i+1}^t(1-\alpha_j) \\
        &=\frac{H+1}{H+i}\bigg(1+\frac{i}{H+i+1}+\frac{i}{H+i+1}\frac{i+1}{H+i+2}+...\bigg)
    \end{split}
\end{equation*}
To simplify the last equality, we conjecture the following identity and prove it by induction:
\begin{equation*}
    \frac{n}{k}=1+\frac{n-k}{n+1}+\frac{n-k}{n+1}\frac{n-k+1}{n+2}+...
\end{equation*}
where $n, k > 0$ and $n\ge k$.

Note that this is equivalent to induction on $\frac{n}{k}-\sum_{i=0}^tx_i=\frac{n-k}{k}\prod_{i=1}^t\frac{n-k+i}{n+i}$. For the base case $t=1$, we get $\frac{n}{k}-1-\frac{n-k}{n+1}=\frac{n-k}{k}-\frac{n-k}{n+1}$ and $\frac{n-k}{k}\frac{n-k+1}{n+1}=\frac{n-k}{k}\bigg(1-\frac{k}{n+1}\bigg)=\frac{n-k}{k}-\frac{n-k}{n+1}$. Assume the induction hypothesis holds for $t=m$ so $\frac{n}{k}-\sum_{i=0}^mx_i=\frac{n-k}{k}\prod_{i=1}^m\frac{n-k+i}{n+i}$. For $t=m+1$, 
\begin{equation*}
    \begin{split}
        \frac{n}{k}-\sum_{i=0}^{m}x_i-x_{m+1}&=\frac{n-k}{k}\prod_{i=1}^m\frac{n-k+i}{n+i}-x_{m+1} \\
        &=\frac{n-k}{k}\prod_{i=1}^m\frac{n-k+i}{n+i}-\prod_{i=1}^{m+1}\frac{n-k+i-1}{n+i} \\
        &=\bigg(\frac{n-k}{k}\prod_{i=1}^m\frac{n-k+i}{n+i}\bigg)\bigg(1-\frac{k}{n+m+1}\bigg) \\
        &=\bigg(\frac{n-k}{k}\prod_{i=1}^m\frac{n-k+i}{n+i}\bigg)\bigg(\frac{n-k+m+1}{n+m+1}\bigg)\\
        &=\frac{n-k}{k}\prod_{i=1}^{m+1}\frac{n-k+i}{n+i}
    \end{split}
\end{equation*}
This finishes the induction. By taking $n=H+i$ and $k=H$, we get $\sum_{t=1}^\infty\alpha_t^i=\frac{H+1}{H+i}\frac{H+i}{H}=\frac{H+1}{H}=1+\frac{1}{H}$. This concludes the proof of Lemma 1.1.

\textbf{Lemma 1.2.} \textit{For any $(x,a,h)\in\cS\times\cA\times[H]$ and episode $k\in[K]$ let $t=N^k_h(x,a)$ and suppose $(x,a)$ was previously taken at step $h$ of episodes $k_1,k_2,...,k_t<k$. Then:}
\begin{equation*}
    (Q_h^k-Q_h^*)(x,a)=\alpha_t^0(H-Q_h^*(x,a))+\sum_{i=1}^t\alpha_t^i\bigg[(V_{h+1}^{k_i}-V_{h+1}^*)(x_{h+1}^{k_i})+[(\hat{\PP}_h^{k_i}-\PP_h)V_{h+1}^*](x,a)+b_i\bigg]
\end{equation*}

\textit{Proof of Lemma 1.2.} Recall that $\sum_{i=1}^t\alpha_t^i=1$ and $[\hat{\PP}_h^{k_i}V_{h+1}](x,a)=V_{h+1}(x_{h+1}^{k_i})$. The Bellman optimality equation is $Q^*_h(x,a)=(r_h+\PP_hV_{h+1}^*)(x,a)$. Then, $\sum_{i=1}^t\alpha_t^ir_h(x,a)=r_h(x,a)\sum_{i=1}^t\alpha_t^i=r_h(x,a)$. Similarly, $[\PP_hV_{h+1}^*](x,a)=[\PP_hV_{h+1}^*](x,a)-[\hat{\PP}_h^{k_i}V_{h+1}^*](x,a)+V_{h+1}^*(x_{h+1}^{k_i})$ and the same trick with $\sum_{i=1}^t\alpha_t^i$ applies here too. Furthermore, $Q_h^*(x,a)=(\alpha_t^0Q_h^*+r_h+\PP_hV_{h+1}^*)(x,a)$ where $\alpha_t^0=
\begin{cases} 0, & \text{$t\ge1$} \\ 1, & \text{$t=0$} \end{cases}$
. This manipulation is valid since $t=1$ represents the start of the episode so $Q_h^*(x,a)$ is technically just defined as itself at $t=0$. By consolidating everything we get: 
\begin{equation}\label{eq:3}
    Q_h^*(x,a)=\alpha_t^0Q_h^*(x,a)+\sum_{i=1}^t\alpha_t^i\bigg[r_h(x,a)+(\PP_h-\hat{\PP}_h^{k_i})V_{h+1}^*(x,a)+V_{h+1}^*(x_{h+1}^{k_i})\bigg]
\end{equation}
We attain Lemma 1.2 by $Q_h^k(x,a)-Q_h^*(x,a)$ where $Q_h^k(x,a)$ comes from \eqref{eq:2} and $Q_h^*(x,a)$ comes from \eqref{eq:3}. This concludes the proof of Lemma 1.2.

\textbf{Lemma 1.3.} \textit{There exists an absolute constant $c>0$ such that, for any $p\in(0,1)$, letting $b_t=c\sqrt{H^3\iota/t}$, we have $\beta_t=2\sum_{i=1}^t\alpha_t^ib_i\le4c\sqrt{H^3\iota/t}$ and, with probability at least $1-p$, the following holds simultaneously $\forall (x,a,h,k)\in\cS\times\cA\times[H]\times[K]:$}
\begin{equation*}
    0\le(Q_h^k-Q_h^*)(x,a)\le\alpha_t^0H+\sum_{i=1}^t\alpha_t^i(V_{h+1}^{k_i}-V_{h+1}^*)(x_{h+1}^{k_i})+\beta_t
\end{equation*}

\textbf{Notation.} The idea behind this lemma is to construct an upper confidence bound on the optimal state-action values, $Q_h^*$ $\forall$ $h \in \{1, 2, \dots, H\}$. Before going into the proof, we first define some notation.

For each state-action-step pair $(x, a, h) \in \cS\times\cA\times[H]$, we denote $k_i$ as the episode in which $(x, a, h)$ occurs for the $i^{th}$ time. Otherwise, $k_i$, $k_{i + 1}$, $\dots$, $k_K = K + 1$ if $(x, a, h)$ only occurs $i - 1$ times over the $K$ episodes. It is important to note that the $K$ episodes are indexed based on the ordering in which they were observed, that is, $k = j$ indicates the $j^{th}$ episode observed. Consequently, $k_i$ is denoted as
\begin{equation*}
    k_i = \begin{cases} \min (k \in [K] \mid \{k > k_{i - 1} \wedge (x_h^k, a_h^k)\} \cup \{K + 1\}), & \quad i \in [K] \\ 0, & \quad i = 0 \end {cases}
\end{equation*}

The aforementioned notation will be utilized for the proofs of this lemma and Theorem 1. 

\textit{Proof of Lemma 1.3.}
For every fixed $(x,a,h)\in\cS\times\cA\times[H]$, let $t = N_h^k(x, a)$, indicating the number of occurrences of $(x, a, h)$ before the start of episode $k$. Moreover, let $\cF_i$ be a $\sigma$-field generated by all random variables up to episode $k_i$, step $h$. In the context of a probability space $(\Omega, \cF, \PP)$, $\{\cF_i\}_{i = 1}^{K}$ is defined as a filtration over $(\Omega, \cF, \PP)$ consisting of an increasing family of sub-$\sigma$-fields \cite{Konstantopoulos2011} of the event space $\cF$, where a $\sigma$-field $\cF_j$ can be interpreted as the accumulative information or collection of events generated from the observation of outcomes from the past episodes $k_1, k_2, \dots k_{j-1}$ and the current episode $k_j$. Note that we will only be concerned with episodes for which outcome $(x, a, h)$ occurs by use of an indicator function $\II[k_i \leq K]$ in the latter half of the proof. 

From the error $[(\hat{\PP}_h^{k_i}-\PP_h)V_{h+1}^*](x,a)$ of the empirical data in Lemma 1.2 along with the predefined notion of $k_i$ and filtration, we now construct the sequence
\begin{equation}\label{eq:4}
    \EE\Big[[(\hat\PP_h^{k_i} - \PP_h)V_{h + 1}^*](x, a) \;|\; \cF_1, \cF_2, \dots, \cF_i \Big] = 0 \quad \forall i \in \{1, 2, \dots, K\}
\end{equation}

The result shown above stems from the fact that taking the expectation of $\hat\PP_h^{k_i}V_{h + 1}^*(x, a)$ conditioned on the past $\sigma$-fields $\cF_1, \cF_2, \dots, \cF_i$ provides knowledge of the probability transition matrix $\PP_h(x^{'}| x,a)$, which implies the following:
\begin{equation*}
    \begin{split}
        \EE\Big[[\hat\PP_h^{k_i}V_{h + 1}^*](x, a) \;|\; \cF_1, \cF_2, \dots, \cF_i \Big] &= \sum_{x' \in \cS} \PP_h(x' \;|\; x, a) \cdot V_{h+1}^*(x') \\
        &= \EE\Big[[\PP_hV_{h + 1}^*](x, a) \;|\; \cF_1, \cF_2, \dots, \cF_i \Big]
    \end{split}
\end{equation*}

Given that we assume the setting to be a tabular episodic finite-horizon MDP, $\cM = (\cS, \cA, H, \PP, r)$, where $|\cS|, |\cA|$, and $H$ are finite with a finite amount of episodes $K$, then
\begin{equation}\label{eq:5}
    \EE\bigg[\:\Big|[(\hat\PP_h^{k_i} - \PP_h)V_{h + 1}^*](x, a)\Big|\:\bigg] < \infty \quad \forall i \in \{1, 2, \dots, K\}
\end{equation}

Since \eqref{eq:4} and \eqref{eq:5} hold true, the sequence of empirical errors $\{[(\hat\PP_h^{k_i} - \PP_h)V_{h + 1}^*](x, a)\}_{i = 1}^{K}$ can be interpreted as a martingale difference sequence (MDS) with respect to the filtration $\{\cF\}_{i = 1}^K$ \cite{MDS}. Therefore, we can use the Azuma-Hoeffding inequality to give a concentration result \cite{williams_1991} for each index in the MDS, i.e., to construct confidence bounds for $Q_h^*$ $\forall$ $h \in \{1, 2, \dots, H\}$. Applying Azuma-Hoeffding and a union bound over all $K$ episodes gives the following:
\begin{equation}\label{eq:6}
    \Bigg| \sum_{i = 1}^{\tau} \alpha_{\tau}^i \cdot \II[k_i \leq K] \cdot [(\hat\PP_h^{k_i} - \PP_h)V_{h + 1}^*](x, a) \Bigg| \leq \frac{cH}{2} \sqrt{\sum_{i = 1}^{\tau} (\alpha_{\tau}^{i})^2 \cdot \iota} \leq c\:\sqrt{\frac{H^3\iota}{\tau}} \quad\quad \forall \tau \in [K]
\end{equation}

for some absolute constant c, with probability at least $1 - \frac{p}{SAH}$. Recall that $\II[k_i \leq K]$ is an indicator function that filters out episodes where $(x, a)$ $was$ $not$ $taken$ $at$ $step$ $h$. To prove the left inequality in \eqref{eq:6}, we consider a previously stated fact that $r_h \in [0, 1]$, implying $Q_h(x, a) \leq H$ and thus $V_h(x) \leq H$ for any $x, a, h$:
\begin{equation}\label{eq:7}
     \begin{split}
        \Bigg| [(\hat\PP_h^{k_i} - \PP_h)V_{h + 1}^*](x, a) \Bigg| \leq H \leq cH \leq \sqrt{2}cH \leq \sqrt{2} \alpha_{\tau}^i cH = c_i\\
        \forall i \in \{1, 2, \dots, K\},\; c > 0 
    \end{split}
\end{equation}

Note that $c_i$ is the symmetric bound on the martingale difference $[(\hat\PP_h^{k_i} - \PP_h)V_{h + 1}^*](x, a)$ and is used for the Azuma-Hoeffding inequality:
\begin{equation}\label{eq:8}
     \begin{split}
        \PP \Bigg[ \; \bigg| \sum_{i = 1}^{\tau} \alpha_{\tau}^i \cdot \II(k_i \leq K) \cdot [(\hat\PP_h^{k_i} - \PP_h)V_{h + 1}^*](x, a) \bigg| \geq \epsilon \; \Bigg] \leq 2\,\exp\bigg(-\frac{2\epsilon^2}{\sum_{i = 1}^K c_i^2}\bigg)
        \end{split}
\end{equation}

whose complementary event is
\begin{equation}\label{eq:9}
    \PP \Bigg[ \; \bigg| \sum_{i = 1}^{\tau} \alpha_{\tau}^i \cdot \II(k_i \leq K) \cdot [(\hat\PP_h^{k_i} - \PP _h)V_{h + 1}^*](x, a) \bigg| \leq \epsilon \; \Bigg] \geq 1 - 2\,\exp\bigg(-\frac{2\epsilon^2}{\sum_{i = 1}^K c_i^2}\bigg)
\end{equation}

To find the proper choice of $\epsilon$, we revisit the bound on the martingale difference: 
\begin{equation}\label{eq:10}
    \begin{split}
        \Bigg| \sum_{i = 1}^{\tau} \alpha_{\tau}^i \cdot \II(k_i \leq K) \cdot [(\hat\PP_h^{k_i} - \PP_h)V_{h + 1}^*](x,a) \Bigg| &\leq \Bigg| \sum_{i = 1}^{\tau} \alpha_{\tau}^i \cdot \II(k_i \leq K) \cdot cH \Bigg| \\
        &= cH\sqrt{\sum_{i = 1}^{\tau} (\alpha_{\tau}^i)^2 \cdot (\II(k_i \leq K))^2} \\
        &\leq cH\sqrt{\sum_{i = 1}^{\tau} (\alpha_{\tau}^i)^2} \leq cH\sqrt{\sum_{i = 1}^{\tau} (\alpha_{\tau}^i)^2 \cdot \iota} = \epsilon
    \end{split}
\end{equation}

With $c_i$ in \eqref{eq:7} and $\epsilon$ in \eqref{eq:10}, we can rewrite the right-hand side of the inequality in \eqref{eq:9} as
\begin{equation}\label{eq:11}
    1 - 2\,\exp\bigg(-\frac{2\epsilon^2}{\sum_{i = 1}^K c_i^2}\bigg) = 1 - \frac{2p}{SAH}
\end{equation}

Therefore, results from \eqref{eq:9} and \eqref{eq:11} indicate an upper bound on the left-hand side of \eqref{eq:6} with a probability of $1 - \frac{2p}{SAH}$. Rescaling $p$ to $\frac{p}{2}$ finishes the proof of the left inequality of \eqref{eq:6}. 

To remove the notation of learning rate as shown on the right-hand side of \eqref{eq:6}, we apply property (b) of Lemma 1.1, which gave an inclusive upper bound of $\frac{2H}{t}$ for $\sum_{i=1}^t(\alpha_t^i)^2$, $\forall t \geq 1$. Making the substitution on the middle term of \eqref{eq:6}, that is, $\frac{cH}{2} \sqrt{\sum_{i = 1}^{\tau} (\alpha_{\tau}^{i})^2 \cdot \iota}$, concludes the proof of \eqref{eq:6}.

Because the inequality in \eqref{eq:6} holds for all fixed $\tau$ $\in$ $[K]$ uniformly, it also holds for $\tau = t = N_h^k(x, a) \leq [K]$. As a result, we can rewrite \eqref{eq:6} in a way that removes the indicator function:
\begin{equation}\label{eq:12}
    \Bigg| \sum_{i = 1}^{t} \alpha_{t}^i \cdot [(\hat\PP_h^{k_i} - \PP_h)V_{h + 1}^*](x, a) \Bigg| \leq c\:\sqrt{\frac{H^3\iota}{t}} \quad where \: t = N_h^k(x,a)
\end{equation}

If we choose the Hoeffding-style bonus $b_t$ to be $c \sqrt{\frac{H^3 \iota}{t}}$ from the equation above, then from property (a) in Lemma 1.1, 
\begin{equation}\label{eq:13}
    \sum_{i=1}^t \alpha_t^ib_i = \sum_{i=1}^t \alpha_t^i\cdot c\sqrt{\frac{H^3\iota}{t}} \in \Bigg[\,c\sqrt{\frac{H^3\iota}{t}}, \: 2c\sqrt{\frac{H^3\iota}{t}}\,\Bigg]
\end{equation}

For notational convenience, we introduce $\frac{\beta}{2} = \sum_{i=1}^t \alpha_t^ib_i$. The final step is putting everything together to yield an upper confidence bound for $Q_h^*$:
\begin{equation*}
    \begin{split}
        (Q_h^k - Q_h^*)(x, a) &\leq \alpha_t^0\cdot (H-Q_h^*(x,a)) + \sum_{i=1}^t\alpha_t^i\cdot \bigg( (V_{h+1}^{k_i} - V_{h+1}^*)(x_{h+1}^{k_i}) + \Big[ (\hat\PP_h^{k_i} - \PP_h) V_{h+1}^* \Big](x_h^{k_i}, a_h^{k_i}) + b_i\bigg) \\
        &\leq \alpha_t^0\cdot (H-Q_h^*(x,a)) + \sum_{i=1}^t\alpha_t^i\cdot \bigg( (V_{h+1}^{k_i} - V_{h+1}^*)(x_{h+1}^{k_i}) + b_i\bigg) + c\sqrt{\frac{H^3\iota}{t}}\\
        &\leq \alpha_t^0\cdot H + \sum_{i=1}^t\alpha_t^i\cdot \bigg( (V_{h+1}^{k_i} - V_{h+1}^*)(x_{h+1}^{k_i}) \bigg) + \beta_t
    \end{split}
\end{equation*}

where the first inequality stems immediately from Lemma 1.2. The right inequality in \eqref{eq:6} is then applied as an inclusive upper bound for the next step. Lastly, the definition of $\beta$ and the fact that $\sum_{i=1}^t \alpha_t^i \leq 1$ are utilized to construct the final inequality, thus completing the proof of Lemma 1.3.

\textbf{Proof of Theorem 1.}

\begin{figure}[!htbp]
  \centering
  \includegraphics[width=0.8\linewidth]{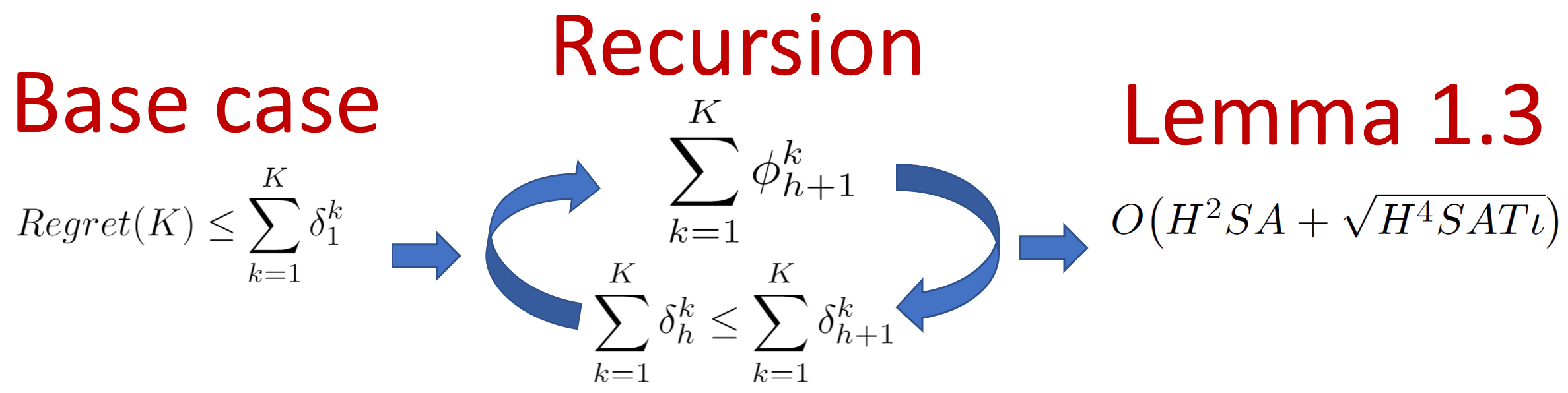}
  \caption{Flowchart for the Proof of Theorem 1.}
  \label{fig:proof_flow}
\end{figure}

The proof of Theorem 1 uses Lemma 1.3 and the Azuma-Hoeffding inequality to produce a recursive formulation for the upper bound of the regret. 
Figure \ref{fig:proof_flow} illustrates the high-level flow of the proof to follow. 

We define $\delta^k_h := (V_{h}^{k}-V_{h}^{\pi_k})(x_{h}^{k})$ and $\phi^k_h := (V_{h}^{k}-V_{h}^{\star})(x_{h}^{k})$. Using Lemma 1.3, the regret can be upper bounded as $Regret(K) \le \sum_{k=1}^{K}\delta^k_1$. 

The main idea is to upper bound $\sum_{k=1}^{K}\delta^k_h$ by the next step $\sum_{k=1}^{K}\delta^k_{h+1}$ which gives a recursive relation for the total regret. For any fixed $(k,h) \in [K] \times [H]$, let $t = N_h^k(x_h^k,a_h^k)$ and suppose $(x_h^k,a_h^k)$ was previously taken at step $h$ of episodes $k_1, k_2,...,k_t < k$. Then, 
\begin{equation} \label{eq:14}
    \begin{split}
        \delta^k_h \leq  (Q^k_h -Q^{\pi_k}_h)(x^k_h, a^k_h) & = (Q^k_h- Q^\star_h)(x^k_h, a^k_h) + (Q^\star_h - Q^{\pi_k}_h)(x^k_h, a^k_h) \\
        & \le \alpha_t^0H + \sum_{i=1}^t\alpha_t^i\phi_{h+1}^{k_i} + \beta_t + [\PP_h(V_{h+1}^\star - V_{h+1}^{\pi_k} )](x^k_h, a^k_h) \\
        & = \alpha_t^0H + \sum_{i=1}^t\alpha_t^i\phi_{h+1}^{k_i} + \beta_t - \phi_{h+1}^k + \delta_{h+1}^k + \epsilon_{h+1}^k 
    \end{split}
\end{equation}
where $\beta_t = 2\sum\alpha_t^ib_i \le \cO(1)\sqrt{H^3\iota/t}$ and $\epsilon_{h+1}^k = [(\PP_h - \hat{\PP}_h^k)(V_{h+1}^\star - V_{h+1}^k )](x_h^k,a_h^k)$ is a martingale difference sequence. Line 1 uses the definition of Q-value function and $V_h^k(x_h^k) \le \max_{a' \in A} Q_h^k(x_h^k, a') = Q_h^k(x_h^k, a_h^k)$. Line 2 follows from Lemma 1.3, the Bellman equation $Q_h^\pi(x,a)=(r_h+\PP_hV_{h+1}^\pi)(x,a)$ and Bellman optimality equation $Q_h^*(x,a)=(r_h+\PP_hV_{h+1}^*)(x,a)$. Finally, Line 3 holds by definition of $\delta_{h+1}^k - \phi_{h+1}^k = (V_{h+1}^{\star} - V_{h+1}^{\pi_k})(x_{h+1}^k)$.

Now, we use \eqref{eq:14} to compute $\sum_{k=1}^K\delta_h^k$. Hence, we get: 
\begin{equation}\label{eq:15}
    \sum_{k=1}^K\delta_h^k \le \sum_{k=1}^K\alpha_t^0H + \sum_{k=1}^K\sum_{i=1}^{n_h^k}\alpha_{n_h^k}^i\phi_{h+1}^{k_i(x_h^k,a_h^k)}+\sum_{k=1}^K\delta_{h+1}^k + \sum_{k=1}^K(\beta_{n_h^k} + \epsilon_{h+1}^k)
\end{equation}

Let $n_h^k = t=N_h^k(x_h^k, a_h^k)$. The first term of \eqref{eq:15} is $\sum_{k=1}^K\alpha_{n_h^k}^0H = \sum_{k=1}^KH \cdot \II[n_h^k = 0] \le SAH$. The equality follows from $\alpha_t^0=
\begin{cases} 0, & \text{$t\ge1$} \\ 1, & \text{$t=0$} \end{cases}$. The inequality stems from the fact that, in the worst case, $n_h^k=0$ for all state-action pairs $(x,a)$ which results in an upper bound of $SAH$.

Next, we bound the second term of \eqref{eq:15}: $\sum_{k=1}^K\sum_{i=1}^{n_h^k}\alpha_{n_h^k}^i\phi_{h+1}^{k_i(x_h^k,a_h^k)}$ where $k_i(x_h^k,a_h^k)$ is the episode in which $(x_h^k, a_h^k)$ was taken at step $h$ for the $i$th time. We first reorder the $\alpha^i_{n_h^k}$ and $\phi_{h+1}^{k_i(x_h^k,a_h^k)}$ terms. Note that $n_h^k = n_h^{k'} + j$ where $j=1,2,...$ is the $j$th time $\phi_{h+1}^{k'}$ appears in the summand due to the fact that $\forall$ $k' \in [K]$, the term $\phi_{h+1}^{k'}$ appears in the summand with $k > k'$ if and only if $(x_h^k, a_h^k) = (x_h^{k'}, a_h^{k'})$.This results in the following simplification:
\begin{equation*}
     \sum_{k=1}^K\sum_{i=1}^{n_h^k}\alpha_{n_h^k}^i\phi_{h+1}^{k_i(x_h^k,a_h^k)} \le \sum_{k=1}^K\phi_{h+1}^{k'} \sum_{t=n_h^{k'}+1}\alpha_t^{n_h^{k'}} \le \bigg(1 + \frac{1}{H}\bigg)\sum_{k=1}^K\phi_{h+1}^k
\end{equation*}
where the first inequality uses the reasoning above and the final inequality uses property (c) of Lemma 1.1. Plugging the above inequalities into \eqref{eq:15} results in:
\begin{equation} \label{eq:16}
    \begin{split}
        \sum_{k=1}^K\delta_h^k &\le SAH + \bigg(1 + \frac{1}{H}\bigg)\sum_{k=1}^K\phi_{h+1}^k - \sum_{k=1}^K\phi_{h+1}^k + \sum_{k=1}^K\delta_{h+1}^k + \sum_{k=1}^K(\beta_{n_h^k} + \epsilon_{h+1}^k)\\
        & = SAH + \frac{1}{H}\sum_{k=1}^K\phi_{h+1}^k+\sum_{k=1}^K\delta_{h+1}^k + \sum_{k=1}^K(\beta_{n_h^k} + \epsilon_{h+1}^k)\\
        & \le SAH + \bigg(1+\frac{1}{H}\bigg)\sum_{k=1}^K\delta_{h+1}^k + \sum_{k=1}^K(\beta_{n_h^k} + \epsilon_{h+1}^k)
    \end{split}
\end{equation} 
where the last inequality is true because $\phi_{h+1}^k\le\delta_{h+1}^k$ since $V^*\ge V^\pi_k$.

Inequality \eqref{eq:16} recursively upper bounds $\sum_{k=1}^{K}\delta^k_h$ by $\sum_{k=1}^{K}\delta^k_{h+1}$. Applying recursion for steps $h \in \{1,2,....,H\}$ and using $\delta_{H+1}^K = 0$ (the algorithm reaches the terminal state so $V_{H+1}^K=V_{H+1}^{\pi_K}=0$) gives:
\begin{equation} \label{eq:17}
    \begin{split}
        \sum_{k=1}^K\delta_1^k &\le SAH + \bigg(1+\frac{1}{H}\bigg)\sum_{k=1}^K\delta_{2}^k + \sum_{k=1}^K(\beta_{n_h^k} + \epsilon_{h+1}^k) \\
        &\le SAH + \bigg(1+\frac{1}{H}\bigg)\bigg[SAH + \bigg(1+\frac{1}{H}\bigg)\sum_{k=1}^K\delta_{3}^k + \sum_{k=1}^K(\beta_{n_h^k} + \epsilon_{h+1}^k)\bigg]+\sum_{k=1}^K(\beta_{n_h^k} + \epsilon_{h+1}^k)\\
        &= SAH+\bigg(1+\frac{1}{H}\bigg)SAH+\bigg(1+\frac{1}{H}\bigg)^2SAH+\dots+\bigg(1+\frac{1}{H}\bigg)^{H-1}SAH \\ 
        & \qquad +\cO\bigg(\sum_{h=1}^H\sum_{k=1}^K(\beta_{n_h^k} + \epsilon_{h+1}^k)\bigg)\\
        &= \cO\bigg(H^2SA + \sum_{h=1}^H\sum_{k=1}^K(\beta_{n_h^k} + \epsilon_{h+1}^k)\bigg)    
    \end{split}
\end{equation}
Overall, we achieve $\sum_{k=1}^K\delta_1^k\le\cO\bigg(H^2SA + \sum_{h=1}^H\sum_{k=1}^K(\beta_{n_h^k} + \epsilon_{h+1}^k)\bigg)$ from \eqref{eq:17}.

By definition of $\beta$, we have $\sum_{k=1}^K\beta_{n_h^k}\le\cO(1)\cdot\sum_{k=1}^K\sqrt{\frac{H^3\iota}{n_h^k}}$. Applying the pigeon-hole principle to the inequality would mean the following: Suppose we play $1/\sqrt{n}$ at a state-action pair $(x,a)$. If we visit $(x,a)$ again, then we only need to play $1/\sqrt{n+1}$ since we cannot include $1/\sqrt{n}$ twice in the summation for the same $(x,a)$. Thus for every $(x,a)$, we have $\sum_{n=1}^{N_h^K(x,a)}\sqrt{\frac{1}{n}}$. Hence we get:
\begin{equation*}
    \sum_{k=1}^{K}\beta_{n_h^k} \le \cO(1)\cdot\sum_{k=1}^K\sqrt{\frac{H^3\iota}{n_h^k}} = \cO(1)\cdot \sum_{x,a}\sum_{n=1}^{N_h^K(x,a)}\sqrt{\frac{H^3\iota}{n}}
\end{equation*}

Note that $\sum_{x,a}N_h^K(x,a)=K$ because we are summing all occurrences of state-action pairs that occur at step $h$ over all episodes. Since there are $K$ episodes, there are $K$ occurrences of state-action pairs occurring at step $h$.

Now, we have $\sqrt{H^3\iota}\sum_{x,a}\sum_{n=1}^{N_h^K}\frac{1}{\sqrt{n}}\le\sqrt{H^3\iota}\sum_{x,a}\sqrt{N_h^K}=\sqrt{H^3\iota}\boldsymbol{1}^Tv$ where $v=[N_h^K(x_1,a_1),N_h^K(x_2,a_2),\dots,N_h^K(x_{SA},a_{SA})]^T$. Using the Cauchy-Schwarz inequality, we get $\sqrt{H^3\iota}\boldsymbol{1}^Tv\le\sqrt{H^3\iota}\sqrt{ SA\sum_{x,a}N_h^K}=\sqrt{H^3SAK\iota}=\sqrt{H^2SAT\iota}$ by realizing that $T=KH$. Consolidating everything in one place, we get the following:
\begin{equation} \label{eq:18}
\sum_{k=1}^{K}\beta_{n_h^k} \le \cO(1)\cdot\sum_{k=1}^K\sqrt{\frac{H^3\iota}{n_h^k}}=\cO(1)\cdot \sum_{x,a}\sum_{n=1}^{N_h^K(x,a)}\sqrt{\frac{H^3\iota}{n}} \le \cO(H^3SAK\iota)=\cO(\sqrt{H^2SAT\iota})
\end{equation}

By the Azuma-Hoeffding inequality, with probability $1-p$, we get:
\begin{equation} \label{eq:19}
\bigg|\sum_{h=1}^H\sum_{k=1}^K\epsilon_{h+1}^k\bigg| = \bigg|\sum_{h=1}^H\sum_{k=1}^K[(\PP_h- \hat{\PP}_h^k)(V_{h+1}^\star) - V_{h+1}^k)](x_h^k,a_h^k)\bigg| \le cH\sqrt{T\iota}    
\end{equation}

Substituting \eqref{eq:18} and \eqref{eq:19} in \eqref{eq:17} gives the following with probability $1-p$ :
\begin{equation*}
    \begin{split}
        \sum_{k=1}^K\delta_1^k & \le \cO(H^2SA+H\sqrt{H^2SAT\iota} + cH\sqrt{T\iota}) \\
        &=\cO(H^2SA+\sqrt{H^4SAT\iota}+c\sqrt{H^2T\iota})\\
        &=\cO(H^2SA + \sqrt{H^4SAT\iota})  
    \end{split}
\end{equation*}
where the final equality is valid since $c\sqrt{H^2T\iota}$ is the smallest of the three terms. This concludes the proof of Theorem 1.
\section*{Conclusion}

In this paper, we showed that a subset of model-free reinforcement learning algorithms can be made sample efficient. Specifically, we proved that, in an episodic setting, Q-learning with UCB-Hoeffding exploration strategy achieves a regret of $\cO(\sqrt{H^4SAT\iota})$. This is the first time a regret analysis features a $\sqrt{T}$ factor for model-free algorithms that do not require access to a "simulator". Thus, the key takeaways from the paper are:
\begin{itemize}
    \item Use UCB exploration over $\varepsilon$-greedy in the model-free setting for better treatment of uncertainties in different states and actions.
    \item Use dynamic learning rates $\alpha_t = \cO(H/t)$ such as $\frac{H+1}{H+t}$ instead of the commonly used $1/t$ for updates at time step $t$. This applies more weight to more recent updates and is critical for sample-efficiency guarantees.
\end{itemize}
We can build upon our current work by examining and unfolding the proof of Q-learning with the more sophisticated UCB-Berstein exploration strategy. Lastly, we can attempt to apply the theoretical framework used in this paper to analyze the pairing of Q-learning with another kind of exploration strategy, such as optimistic initial values.

\newpage
\bibliographystyle{unsrt}
\bibliography{main}

\end{document}